%% file: neurips_2026.tex
\documentclass{article}

 \usepackage[main, final]{neurips_2026}

\usepackage[utf8]{inputenc} 
\usepackage[T1]{fontenc}    
\usepackage{hyperref}       
\usepackage{url}            
\usepackage{booktabs}       
\usepackage{amsfonts}       
\usepackage{nicefrac}       
\usepackage{microtype}      
\usepackage{xcolor}         
\usepackage{algorithm}
\usepackage{algpseudocode}

\usepackage{fancyhdr} 
\usepackage{amsmath}
\input{math_commands.tex}

\usepackage{amssymb}

\usepackage{enumitem}
\usepackage{hyperref}
\usepackage{url}
\usepackage{booktabs}
\usepackage{graphicx}
\usepackage[capitalise]{cleveref}
\usepackage{wrapfig} 

\usepackage{makecell}
\usepackage{colortbl}
\usepackage{multirow}
\usepackage{svg}
\usepackage[dvipsnames]{xcolor}
\usepackage{listings}
\usepackage{amssymb}
\usepackage{nicematrix}

\usepackage{tikz}
\usetikzlibrary{arrows.meta, positioning}
\usepackage[dvipsnames,table,xcdraw]{xcolor}

\usepackage{xcolor}

\usepackage{amsmath}
\usepackage{dsfont}
\definecolor{casegray}{gray}{0.96}

\usepackage[most]{tcolorbox}
\usepackage{xcolor}
\usepackage{enumitem}
\usepackage{amsmath}

\usepackage{adjustbox}

\title{Masked Diffusion Language Agents for Tool-Integrated Chemical Reasoning}

\author{Mengdi Liu\textsuperscript{1,2}\thanks{Equal Contribution. }~, 
Chenghao Jia\textsuperscript{1,2*}, 
Hong Chang\textsuperscript{1,2}\thanks{Corresponding author. }~, 
Shiguang Shan\textsuperscript{1,2} \\
\textsuperscript{1} State Key Laboratory of AI Safety, Institute of Computing Technology, CAS, China\\
\textsuperscript{2} University of Chinese Academy of Sciences (CAS), China\\
\texttt{\{liumengdi23z, jiachenghao25z, changhong, sgshan\}@ict.ac.cn} \\
}

\begin{document}

\maketitle

\begin{abstract}
  
  Designing tool-integrated reasoning agent systems for complex chemical tasks remains a fundamental challenge. Although recent chemical agents have demonstrated promising performance through the integration of external tools, they are still built on the autoregressive paradigm, which is inherently constrained by causal attention and left-to-right generation. As a result, they often struggle with bidirectional molecular understanding, global coordinated tool planning and multi-source evidence integration. To address this challenge, we introduce \textsc{ChemDiffAgent}, the first masked diffusion language agent for tool-integrated chemical reasoning. It reformulates chemical reasoning as iterative denoising over interleaved reasoning and tool-use trajectories, enabling tool-use decisions to be jointly determined under full-trajectory context and thereby better aligning the agent's decision boundary with its knowledge boundary. We further analyze how this paradigm benefits agentic reasoning through harder infilling subproblems during training and flexible arbitrary-order decoding during inference. Then we develop a two-stage post-training framework, comprising agentic supervised fine-tuning and variance-reduced preference optimization, to enhance tool-use and reasoning capabilities. We also construct a comprehensive benchmark covering six chemistry tasks across both single-turn and multi-turn tool-use scenarios. Under matched budgets and evaluation, \textsc{ChemDiffAgent} consistently outperforms autoregressive agents, achieving an overall improvement of approximately 20\%. Notably, our 8B model achieves performance comparable to, in some cases surpassing, significantly larger frontier models, such as GPT 5.5 and Claude Opus 4.7. Code is available at \url{https://github.com/bilemond/ChemDiffAgent}.
  
\end{abstract}

\input{./section/Introduction.tex}

\input{./section/Related_Works.tex}
\input{./section/Preliminaries.tex}

\input{./section/Method.tex}

\input{./section/Experiments.tex}

\input{./section/Conclusion.tex}

\bibliographystyle{plain}  
\bibliography{refs}        


\appendix

\input{./section/Appendix.tex}




\newpage
\input{checklist.tex}

\end{document}

%% file: math_commands.tex
\usepackage{amsmath,amsfonts,bm}

\def\eqref#1{equation~\ref{#1}}

\def\1{\bm{1}}

\DeclareMathAlphabet{\mathsfit}{\encodingdefault}{\sfdefault}{m}{sl}
\SetMathAlphabet{\mathsfit}{bold}{\encodingdefault}{\sfdefault}{bx}{n}



%% file: section/Introduction.tex
\section{Introduction}

With the rapid development of reasoning capabilities \cite{2024_Jaech, teamqwq, 2025_Guo, 2024_Luo, 2025_Guan, liu2026protinvtree, jia2026mast0}, large language models (LLMs) have shown growing potential for solving complex chemistry tasks \cite{wang2024survey,ramos2024areview, davis2023testing, boiko2023autonomous, janakarajan2023language}.
However, chemical reasoning often goes beyond parametric knowledge, requiring an agent to interpret domain-specific modalities, orchestrate tool invocations with valid arguments, and synthesize the returned response into subsequent reasoning steps.
This process, which we refer to as \textit{\textbf{tool-integrated chemical reasoning}}, tightly couples internal reasoning with external tool use, enabling agents to systematically coordinate introspection and interaction.
In this sense, an effective chemistry agent should align its tool-use decision boundary with its actual knowledge boundary: recognizing the limits of its parametric knowledge, judiciously deciding when and which tool to invoke, and reconciling potentially conflicting evidence from multiple sources.

\begin{figure*}[tb]
    \centering
	\includegraphics[width=1\textwidth]{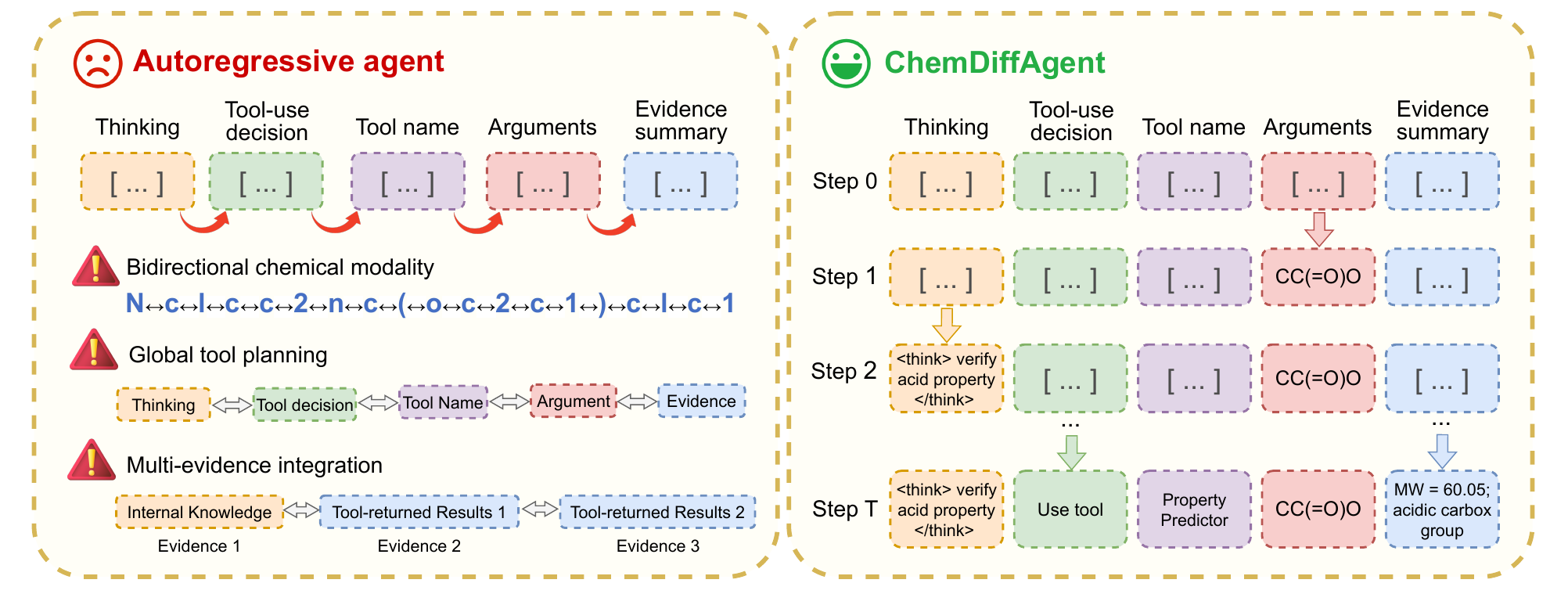}
    \vspace{-0.5cm}
	\caption{Compared with autoregressive agents that commit to tool-use trajectories sequentially, ChemDiffAgent jointly refines chemical modality thinking, tool-use decisions, arguments, and evidences to better handle bidirectional chemical modalities and globally coordinated tool use.}
	\label{fig: illus}
    \vspace{-0.5cm}
\end{figure*}

Recently, chemistry agents have demonstrated promising advances by integrating external tools such as molecular property predictors, reaction engines and cheminformatics libraries \cite{boiko2023autonomous, janakarajan2023language, m2024augmenting, mcnaughton2024cactus}.
However, they rely on the autoregressive (AR) paradigm, where the unidirectional context modeling and left-to-right generation are  inherently misaligned with the nature of chemical reasoning (as shown in Fig. \ref{fig: illus}).
First, chemistry-specific modalities such as SMILES are \emph{bidirectional}, with \emph{complex dependencies} among tokens at different positions.
Second, 
tool-integrated chemical reasoning is \emph{globally coupled}, as the choice of tool, its arguments, and the reasoning goal mutually depend on each other.
Third, complex chemical tasks frequently require integrating \emph{multiple sources of evidences} into a single coherent answer.
Nevertheless, autoregressive agents are inherently constrained by strict left-to-right token commitment, which precludes joint coordination across the sequence and consequently leads to locally biased tool selection and unstable multi-step execution.

To address these limitations, we introduce \textsc{ChemDiffAgent}, the first masked diffusion language agent for tool-integrated chemical reasoning. It reformulates chemical reasoning as iterative denoising over interleaved reasoning and tool-use trajectories.
This formulation allows the agent to jointly determine intermediate thinking, tool-use decisions, tool names, arguments, and multi-evidence within each block under bidirectional context.
As a result, it enables more reliable and flexible chemical reasoning through a unified denoising formulation, rather than being committed prematurely.

We further analyze why this paradigm is well suited for agentic chemical reasoning from both training and inference perspectives.
During agentic training, \textsc{ChemDiffAgent} exposes the model to harder infilling subproblems over interleaved reasoning and tool-use trajectories.
During inference, the arbitrary-order decoding within each block provides greater flexibility for coordinating interdependent decisions, such as selecting a tool, filling its arguments and integrating evidence before committing to the next action.
Building on this analysis, we develop a two-stage post-training framework comprising agentic supervised fine-tuning (SFT) and variance-reduced preference optimization (VRPO) to enhance tool-use decision-making and chemical reasoning capabilities.

We also construct a comprehensive benchmark covering six chemistry tasks across single-turn and multi-turn tool-use scenarios. Under matched budgets and constraints, \textsc{ChemDiffAgent} consistently outperforms autoregressive agents, achieving an overall improvement of approximately 20\%. Notably, our 8B model achieves performance comparable to, and in some cases surpassing, substantially larger frontier models such as GPT-5.5 and Claude Opus 4.7. These results suggest that masked diffusion provides a promising foundation for building more reliable, globally coordinated, and tool-aware agents for complex scientific reasoning.

In summary, our contributions are as follows: 
\begin{itemize}[leftmargin=5.5mm]
    \item We introduce \textsc{ChemDiffAgent}, the first masked diffusion language agent for tool-integrated chemical reasoning. It formulates chemical reasoning as an iterative denoising process over interleaved reasoning and tool-use trajectories, enabling global attention and coordination.
    \item We provide a theoretical analysis revealing why the diffusion paradigm offers advantages over autoregressive generation in tool-integrated chemical reasoning. Building on this analysis, we develop a two-stage post-training framework, comprising Agentic SFT and Agentic VRPO.
    \item We construct a comprehensive benchmark covering six chemistry tasks spanning single-turn and multi-turn tool-use scenarios. 
    Experiments demonstrate that \textsc{ChemDiffAgent} substantially outperforms autoregressive agents and achieves performance comparable to, or even surpassing, larger frontier models such as GPT-5.5 and Claude Opus 4.7.
\end{itemize}

%% file: section/Related_Works.tex
\section{Related Work}
\subsection{Chemistry Agents}

Recent research has demonstrated promising advances in integrating LLMs with external chemistry tools \cite{wang2024survey,ramos2024areview, davis2023testing, boiko2023autonomous, janakarajan2023language} to overcome the unreliability of purely parametric knowledge on domain tasks.
Early efforts, such as StructChem \cite{ouyang2024structured}, broke down chemical reasoning into structured phases, including formula generation and step-by-step reasoning.
ChemCrow \cite{m2024augmenting}, ChemToolAgent \cite{yu2025tooling}, and others \cite{ma-etal-2024-sciagent,mcnaughton2024cactus} further enhanced LLMs by integrating expert-designed tools for tasks like molecular design and drug development.
To enable better \textit{\textbf{tool planning and execution}}, CheMatAgent \cite{wu2025chematagent} presents the Hierarchical Evolutionary MCTS framework, while SciToolAgent \cite{ding2025scitoolagent} leverages a scientific tool knowledge graph that encodes relationships among hundreds of tools.
To solve complex real-world problems, \textbf{\textit{multi-agent systems}} such as MT-Mol~\cite{kim2025mt}, and ChemMAS~\cite{yang2025multi} decompose chemical workflows across role-specialized agents.
Additionally, \textbf{\textit{self-evolving agents}} including ChemAgent~\cite{tang2025chemagent} and S1-NexusAgent~\cite{team2026s1} accumulate experience through self-updating memory libraries.
Despite great progress, these methods rely on autoregressive models, which lack a global understanding of chemical sequences and are constrained by their sequential generation process.
To the best of our knowledge, we are the first to introduce a dLLM chemistry agent, \textit{enabling bidirectional chemical modality understanding, global tool planning and multi-evidence integration}.


\subsection{Diffusion Large Language Models and DLLM Agents}

\paragraph{Diffusion Language Models.} Inspired by discrete diffusion models~\cite{austin2021structured,campbell2022continuous,meng2023concrete,lou2023discrete,zhao2024improving}, dLLMs have emerged as a promising alternative to AR models (ARMs). LLaDA~\cite{LLaDA}, an 8B diffusion language model trained from scratch, achieves performance competitive with LLaMA3-8B. Dream7B~\cite{dream7b} introduces a comprehensive training framework that leverages AR-based LLM initialization and context-adaptive noise scheduling to scale diffusion language models. More recently, industrial efforts such as Gemini Diffusion~\cite{geminidiffusion}, Mercury~\cite{inception2025mercury}, and Seed-Diffusion~\cite{seeddiffusionlargescalediffusion} have further scaled dLLMs and demonstrated their potential for efficient inference.

\paragraph{Block Diffusion Language Models.} Furthermore, hybrid architectures such as Block Diffusion Language Models~(BDLMs)~\cite{BDLM, SDAR, Fastdllmv2, LLaDA2} have become a significant research focus. BDLMs employ attention mechanisms combining intra-block bidirection with inter-block causal. This architecture natively supports KV Cache and variable-length text generation and keeps the non-autoregressive generation ability within each block, allowing parallel decoding in arbitrary orders. SDAR~\cite{SDAR}, a series of BDLMs ranging from 1.7B to 8B parameters, has demonstrated general-purpose capabilities comparable to the latest open-source ARMs.

\paragraph{DLLM Agents.}

Recent works have started to investigate diffusion large language models (dLLMs) as agent backbones. 
DLLM-Agent~\cite{zhen2026dllm} reports over 30\% speedup with fewer interaction rounds at comparable accuracy on BrowseComp-zh and web open-ended tasks compared to AR backbones.
DLLM-Searcher~\cite{zhao2026dllm} combines agentic post-training with a Parallel-Reasoning-and-Acting paradigm that overlaps decoding and tool-waiting, matching mainstream search agents with a 15\% speedup. 
However, these works are confined to web searches. To our knowledge, our work is the first to explore diffusion language agents for scientific reasoning, with a focus on \textit{complex chemical reasoning scenarios involving multi-step, multi-modal, and multi-tool coordination}.





%% file: section/Preliminaries.tex
\section{Preliminaries}

\paragraph{Diffusion Large Language Models.}
Diffusion Large Language Models (dLLMs), or more specifically Masked Diffusion Models (MDMs), define a forward process that gradually corrupts the clean input by replacing tokens with the mask token $\text{M}$. 
Given the prompt $x$ and the clean completions $y$, the forward process $q_t(y_t|y, x)$  at time $t$ is defined as follows: 
\begin{equation}
    q_t(y_t|y,x) = \prod_{i=1}^L q_t(y_t^i|y^i,x) \quad \text{where} \quad q_t(y_t^i|y^i,x)= \begin{cases}
1-t, & y_t^i = y^i, \\
t, & y_t^i = \text{M}.
\end{cases}
\label{eq:de}
\end{equation}

In the reverse process, the model predicts the original values of the masked tokens to compute the reverse probability $p_\theta(\cdot |y_t,x)$. To learn this distribution effectively, dLLMs typically adopts the Evidence Lower Bound (ELBO) $\mathcal{L}_\theta(y | x)$ as a surrogate objective to approximate the exact log-likelihood
$\log \pi_\theta(y|x)$ ~\cite{ou2024your,shi2024simplified,sahoo2024simple}:
\begin{equation}
  \mathcal{L}_\theta(y|x)  \triangleq   \mathbb{E}_{t \sim \mathcal{U}[0,1]} \mathbb{E}_{y_t \sim q_t(y_t |  y, x)} \left[ \frac{1}{t} \sum_{i=1}^{L} \mathbf{1}[y_t^i = \text{M}] \log p_{\theta}(y^i | y_t, x) \right]   \leq  \log \pi_\theta(y|x). 
\label{eq:t-dce-v1}
\end{equation}

For BDLMs, the input $y$ is partitioned into $K$ continuous blocks  $[y^1, \dots,y^K]$, each of length $B$. The corresponding ELBO of is defined as:
\begin{align}
    \label{eq:elbo_split}
    \mathcal{L}^{block}_\theta(y|x)
    \triangleq
    \mathbb{E}_{t \sim \mathcal{U}[0,1]} 
    \mathbb{E}_{y_t \sim q_t(y_t | y, x)}
    \left[
    \frac{1}{t}
    \sum_{k=1}^{K} \sum_{i=1}^{B}
    \mathbf{1}[y_t^{k,i} = \text{M}]
    \log p_{\theta}(y^{k,i} | y_t, y^{<k}, x)
    \right].
\end{align}

%% file: section/Method.tex
\begin{figure*}[tb]
    \centering
	\includegraphics[width=1\textwidth]{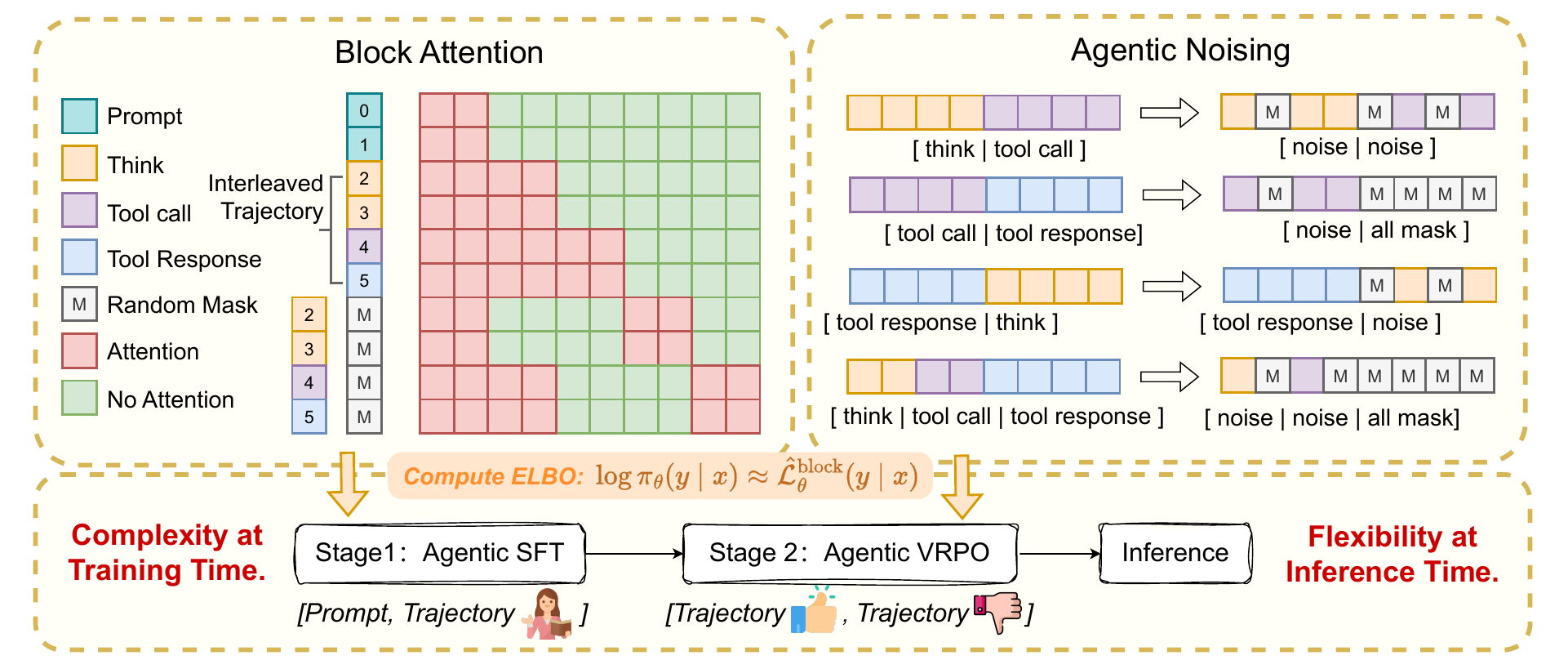}
    \vspace{-0.5cm}
	\caption{Overview of \textsc{ChemDiffAgent}. In training, both Agentic SFT and Agentic VRPO use Block Attention and Agentic Noising to compute the Agentic ELBO. It learns from \textbf{\textit{complex}} block-infilling tasks at training time and enables \textbf{\textit{flexible}} order-agnostic reasoning at inference time.}
	\label{fig: method}
\end{figure*}

\section{\textsc{ChemDiffAgent}}
In this section, we propose \textsc{ChemDiffAgent}, a masked diffusion language agent for tool-integrated chemical reasoning. The overview framework is shown in Figure \ref{fig: method}.

\subsection{Theoretical Foundations of Flexible Tool Reasoning}

\vspace{-0.5em}\paragraph{Iterative Denoising over Interleaved Trajectory.} Given a prompt $x$, \textsc{ChemDiffAgent} produces an interleaved
trajectory $\mathcal{H}=(r_1,c_1,o_1,\ldots,r_T,c_T,o_T,a)$,
where $r_t$ is a reasoning segment, $c_t=(\tau_t,\alpha_t)$
a call to tool $\tau_t$ with arguments $\alpha_t$,
$o_t=\tau_t(\alpha_t)$ its response, and $a$ the final answer.
Unlike autoregressive agents that commit to each token from left to right, \textsc{ChemDiffAgent} models each reasoning--action trajectory as a denoising process over masked blocks. This allows tool names, arguments, control decisions, and reasoning processes to be jointly refined under full block context.


\paragraph{Training Objective Design for Flexible Infilling.}
DLLM agents learn to reconstruct masked content
from partially observed contexts.
This exposes the model to different combinations of available
information rather than a single prefix-based prediction pattern. To characterize this supervision, partition a clean completion $y$
into $K$ contiguous blocks $[y^1,\ldots,y^K]$, where
$y^k=(y^{k,1},\ldots,y^{k,B})$ and $L=KB$.
For a non-empty block set $A\subseteq[K]$, let $y[A]$ replace
all tokens in blocks indexed by $A$ with the mask token $\mathrm{M}$.
Under uniform sampling of the number of masked blocks and of the
block subset conditional on its size, the reconstruction objective is
\begin{equation}
\label{eq:bdlm_block_loss}
\mathcal{L}^{\mathrm{block}}_\theta
=
\mathbb{E}_{(x,y)\sim p_{\mathrm{data}}}
\left[
\sum_{\emptyset\neq A\subseteq[K]}
\frac{
\sum_{k\in A}\sum_{i=1}^{B}
\log p_\theta(y^{k,i}\mid y[A],y^{<k},x)
}{
K\binom{K}{|A|}|A|B
}
\right].
\end{equation}

Here, $p_{\mathrm{data}}$ is the data distribution and $p_\theta$
the model's conditional distribution.
By optimizing $\mathcal{L}^{\mathrm{block}}_\theta$, the model learns to reconstruct missing tokens conditioned on arbitrary visible context combinations. Crucially, this objective directly leads to flexible inference: because $p_\theta$ learns valid posteriors under any mask pattern $A$, it establishes the foundational capacity to evaluate and decode along arbitrary trajectory paths during generation. The detailed derivation is provided in Appendix~\ref{app:block_infilling}.

\paragraph{Adaptive path advantage under non-uniform subproblem difficulty.}
Different parts of a reasoning problem can have different conditional
difficulties, and resolving one part may simplify another.
Let $C=(X,Y^{<k})$ denote the preceding context,
$U=(U_1,\ldots,U_B)$ the current block, and
$S_t=\{i_1,\ldots,i_{t-1}\}$ the positions resolved along a path
$\pi=(i_1,\ldots,i_B)$.
The prediction difficulty at each position, measured by conditional
cross-entropy, consists of conditional entropy and model approximation
error. Summing along the path gives the expected negative log-likelihood
\begin{equation}
\label{eq:adaptive_path_main}
\mathcal{R}_\theta(\pi)
=
H(U\mid C)
+
\mathbb{E}_{p_{\mathrm{data}}}
\left[
\sum_{t=1}^{B}
\delta_\theta(i_t\mid C,U_{S_t})
\right],
\end{equation}

where $H$ denotes conditional entropy and
$\delta_\theta(i\mid C,U_S)
=
D_{\mathrm{KL}}\!\left(
p_{\mathrm{data}}(U_i\mid C,U_S)
\Vert p_\theta(U_i\mid C,U_S)
\right)$
is the model's approximation error.
The accumulated
approximation error depends on the intermediate contexts.
Since the admissible paths $\Pi_{\mathrm{flex}}$
include $\pi_{\mathrm{AR}}=(1,\ldots,B)$,
$\inf_{\pi\in\Pi_{\mathrm{flex}}}\mathcal{R}_\theta(\pi)
\leq\mathcal{R}_\theta(\pi_{\mathrm{AR}})$
for a fixed predictor.
Flexible ordering can therefore exploit differences in subproblem
difficulty to seek a path with smaller accumulated error.
The detailed derivation is provided in Appendix~\ref{app:adaptive_path}.

\paragraph{Strict ordering advantage under non-uniform decision risk.}
Beyond prediction difficulty, errors at different positions can have
different downstream consequences.
Given context $C$, let $Z$ be an earlier high-risk decision in the
AR order and $R$ a later lower-risk decision in the same block that
provides information about $Z$.
Let $w_Z>w_R>0$ be their fixed downstream error-cost weights.
For Bayes-optimal conditional predictions, the risk-weighted log losses
are $\mathcal{J}_{\mathrm{AR}}
=w_ZH(Z\mid C)+w_RH(R\mid C,Z)$ and
$\mathcal{J}_{R\rightarrow Z}
=w_RH(R\mid C)+w_ZH(Z\mid C,R)$.
Their difference satisfies

\begin{equation}
\label{eq:risk_ordering_main}
\mathcal{J}_{\mathrm{AR}}
-
\mathcal{J}_{R\rightarrow Z}
=
(w_Z-w_R)I(Z;R\mid C),
\end{equation}

where $I$ denotes conditional mutual information.
If $I(Z;R\mid C)>0$, resolving $R$
before $Z$ strictly reduces this weighted risk.
Thus, flexible ordering can first resolve lower-risk content and use
more information to guide a higher-risk decision.
The detailed derivation is provided in Appendix~\ref{app:strict_ordering}.

\subsection{Agentic Supervised Fine-Tuning}
\label{subsec:sft}

Directly applying existing dLLM backbones to chemistry-agent tasks exposes two practical limitations.
First, the model must reliably follow the executable tool-calling protocol required by the chemistry tool environment.
Second, it must perform multi-step chemical reasoning rather than produce a single response.
We therefore introduce Agentic SFT to adapt dLLMs to chemistry-oriented tool reasoning.


\paragraph{Teacher Trajectory Construction.}
\label{subsec:tr}
For each query $Q$, we use a high-performance teacher model to generate a candidate chemistry-agent trajectory $\mathcal{H}_{\text{teacher}}$.
The teacher is allowed to select from a small set of relevant chemistry tools rather than the full tool pool, which reduces irrelevant exploration while still requiring tool selection.
We perform trajectory-level rejection sampling and retain a trajectory only if it satisfies three criteria: (1) the final answer is correct; (2) the reasoning process is clear and complete; and (3) every tool selection and argument is correct and strictly follows the required format.
The filtered query--trajectory pairs $(Q, \mathcal{H}_{\text{teacher}})$ are then used as the Agentic SFT training set.

\paragraph{Block Attention for Tool-Use Trajectories.}
To enable efficient training over noised trajectories, we adopt a block-level bidirectional attention scheme tailored to tool-use data, as illustrated in Fig.~\ref{fig: method}. Specifically, we concatenate the query, the clean teacher trajectory, and a noised copy of the trajectory into a single input sequence, $[Q,\mathcal{H}_{\text{teacher}},\mathcal{H}_{\text{noisy}}]$. 
This design allows all noised blocks to be supervised in a single forward pass, conditioned on the corresponding clean trajectory.

\paragraph{Agentic Noising.}
Chemistry trajectories interleave \emph{agent-controlled} decisions, such as \texttt{think} and \texttt{tool\_call} steps, with \emph{external observations} from \texttt{tool\_response} spans. 
Agentic Noising perturbs only agent-controlled regions, leaving tool responses untouched as grounded context. 
To prevent train--inference leakage from intra-block bidirectional attention, we fully mask any \texttt{tool\_response} token
that would otherwise be visible within the same bidirectional block.

\paragraph{Agentic ELBO.}
The training objective should also reflect the fact that only agent-side decisions are supervised.
We therefore modify Eq.~\ref{eq:elbo_split} and define the Agentic ELBO $\hat{\mathcal{L}}^{block}_\theta$ as:
\vspace{+0.1cm}
\begin{align}
\label{eq:sft_loss_aligned} 
\hat{\mathcal{L}}&^{block}_\theta(y \mid x)  \triangleq  \ \mathbb{E}_{t \sim \mathcal{U}[0,1], y_t \sim \hat{q}_t(y_t \mid y, x)}  \\&
\Bigg[ \frac{1}{t} \sum_{k=1}^K\sum_{i=1}^{B} \mathbf{1}\Big[ (y_t^{k,i} = [\text{M}]) \land (y^{k,i} \notin [\text{R}]) \Big] \log p_\theta(y^{k,i} \mid y_t^k,y^{<k}, x) \Bigg], \nonumber
\end{align}

where $[\text{R}]$ denotes the \texttt{tool\_response} regions.
The loss is computed only on positions that are masked and do not belong to external tool observations.
These positions are excluded from the loss because they are not generated by the agent.


\paragraph{Training Objective.}

Instead of optimizing the standard autoregressive negative log-likelihood, we use the proposed Agentic ELBO to approximate $\log \pi_{\theta}(y \mid x)$ under the dLLM objective, where $x$ denotes the prompt and $y$ corresponds to the teacher trajectory $\mathcal{H}_{\text{teacher}}$.
The final SFT loss is:
\vspace{+0.1cm}
\begin{equation}
    \mathcal{L}_{SFT} =  \mathbb{E}_{(x, y) \sim \mathcal{D}} \big[- \hat{\mathcal{L}}^{block}_{\theta}(y \mid x)\big].
\end{equation}

\subsection{Agentic Variance-Reduced Preference Optimization}
\label{subsec:cspo}

Agentic SFT improves the model's capability of executable chemistry tool calling, but it does not explicitly optimize preference between successful and failed multi-step trajectories.
Motivated by prior post-training methods for dLLMs~\cite{VRPO}, we further introduce Agentic VRPO to improve long-horizon chemistry-agent decision making.
The key idea is to contrast a successful trajectory with a failed one under the same query, with both evaluated via the Agentic ELBO.




\paragraph{Variance-Reduced Agentic Scoring.}
For each trajectory $y$, we sample $M$ corrupted contexts through
Agentic Noising.
Let $\xi_j=(t_j,y_{t_j})$ denote the noise time and corrupted
trajectory in the $j$-th sample.
The corresponding agent-only reconstruction score is
\vspace{+0.1cm}
\begin{equation}
\label{eq:cspo_sample_score}
\ell_\theta(y\mid x;\xi_j)
=
\frac{1}{t_j}
\sum_{k=1}^{K}\sum_{i=1}^{B}
\mathbf{1}\!\left[
y_{t_j}^{k,i}=\mathrm{M},\,
(k,i)\notin\mathcal{O}(y)
\right]
\log p_\theta(y^{k,i}\mid y_{t_j}^{k},y^{<k},x),
\end{equation}
\vspace{+0.1cm}
where $\mathcal{O}(y)$ is the set of tool-response positions.
Agentic Noising masks future tool observations,
while the indicator excludes
external observations from scoring. Using the same corrupted input for the policy and fixed reference
model, we estimate their Agentic ELBO difference as
\begin{equation}
\label{eq:cspo_score}
\widehat{\Delta}_M(y\mid x)
=
\frac{1}{M}\sum_{j=1}^{M}
\left[
\ell_\theta(y\mid x;\xi_j)
-
\ell_{\mathrm{ref}}(y\mid x;\xi_j)
\right].
\end{equation}

\paragraph{Training Objective.}
The aggregated scores define the preference objective:
\begin{equation}
\label{eq:cspo_loss}
\mathcal{L}_{\mathrm{VRPO}}(\theta)
=
-\mathbb{E}_{(x,y_w,y_l)\sim\mathcal{D}_{\mathrm{pref}},\,\Xi}
\left[
\log\sigma\left(
\beta\left[
\widehat{\Delta}_M(y_w\mid x)
-
\widehat{\Delta}_M(y_l\mid x)
\right]
\right)
\right],
\end{equation}
where $\Xi$ collects the sampled contexts for both trajectories,
$\sigma$ is the sigmoid function, and $\beta>0$ controls the
preference scale.
Scores are aggregated to reduce reliance on individual context samples.

%% file: section/Experiments.tex
\section{Experiments}
\label{sec:experiments}

\subsection{Dataset and Experimental Setup}
\label{sec:exp_setup}

\vspace{-0.5em}\paragraph{Single-turn dataset construction.}
We first collect chemistry datasets covering six fundamental molecular- and reaction-centered tasks, including composition inference, name conversion, property prediction, retrosynthesis, molecular optimization, and forward synthesis. 
We construct the single-turn split by preserving the original task semantics and using a strong teacher model to generate an execution trajectory. The trajectory construction details are shown in section \ref{subsec:tr}.

\vspace{-0.5em}\paragraph{Multi-turn dataset construction.}
For multi-turn construction, we formulated each instance as a \textit{graph-conditioned tool-use task}.
Inspired by SciToolAgent~\cite{ding2025scitoolagent}, we first build a chemistry tool graph, where nodes represent tools and edges denote the input--output compatibility.
We then perform random walks over the graph to sample subgraphs and use a strong teacher model to compose a query that combines the original task with all sampled tools.
Trajectories are subsequently constructed by executing the tool chain in a forward manner.
We retain a candidate only if its tool-calling trajectory matches the sampled subgraph, followed by manual inspection for query coherence, valid tool use, and answer faithfulness.
The resulting multi-turn examples contain execution-verified final answers and gold tool-calling graph.
Based on the length of the gold calling chain, we partition it into three difficulty levels, namely Easy, Medium, and Hard, corresponding to 2, 3–5, and 6–7 tool-calling rounds respectively.
More construction details are provided in Appendix~\ref{app:benchmark_construction}.

\vspace{-0.5em}\paragraph{Evaluation metrics.}
We report two main metrics: (i) answer correctness, judged by GPT-4o on whether the final response solves the task, and (ii) tool selection accuracy, the fraction of the $m$ expected tools in the gold chain invoked by the model. Full protocols are in Appendix~\ref{app:metrics}.

\begin{table*}[t]
\centering
\caption{Single-turn success rate performance across six chemistry task families: composition inference, name conversion, property prediction, retrosynthesis, molecular optimization, and forward synthesis.
Red numbers denote absolute gains of SDAR-8B+SFT over Qwen3-8B+SFT, and of SDAR-8B+SFT+RL over SDAR-8B+SFT. The \textbf{best} and \underline{second-best} results are highlighted.
}
\vspace{+0.5em}
\label{tab:single_task_family}
\footnotesize
\setlength{\tabcolsep}{5pt}
\renewcommand{\arraystretch}{1.08}
\begin{adjustbox}{max width=\textwidth}
\begin{tabular}{lcccccc|c}
\toprule
Method &
Comp. &
Name Conv. &
Prop. Pred. &
Retro. &
Mol. Opt. &
Fwd. Synth. &
Total \\
\midrule

\rowcolor{gray!12}
\multicolumn{8}{l}{\textit{Strong LLMs with ReAct}}\\
GLM-5.1
& 39.8 & 19.6 & 40.6 & 38.5 & 0.0 & 30.0 & 32.8 \\
Gemini 2.5 Flash
& 77.9 & 40.2 & 67.7 & 0.0 & 79.2 & 75.0 & 59.8 \\
DeepSeek-V3.2
& 83.5 & 71.6 & 90.6 & 50.8 & \underline{88.7} & 78.3 & 77.0 \\
GPT-5.5
& 71.4 & 56.9 & 64.6 & \underline{96.9} & 64.2 & 91.7 & 70.6 \\
Claude Opus 4.7
& 90.9 & 88.2 & 87.5 & \textbf{100.0} & \textbf{100.0} & \textbf{100.0} & \underline{90.9} \\

\midrule

\rowcolor{gray!12}
\multicolumn{8}{l}{\textit{Chemistry agents}}\\
ChemCrow \tiny{(Claude Opus 4.7)}
& 91.2 & \underline{92.2} & 83.3 & 95.4 & \textbf{100.0} & 86.7 & 89.6 \\

ChemCrow \tiny{(DeepSeek-V3.2)}
& 55.0 & 53.2 & 53.0 & 72.6 & 43.1 & 28.3  & 72.4 \\

S1-NexusAgent\tiny{(GPT-5.5)}
& 71.4 & 78.8 & 77.5 & 71.8 & 38.5 & 33.3 & 76.7 \\

\midrule

\rowcolor{gray!12}
\multicolumn{8}{l}{\textit{Autoregressive 8B model}}\\
Qwen3-8B + SFT
& 76.6 & 53.9 & 87.5 & 78.5 & 74.5 & 63.3 & 69.1 \\
\midrule

\rowcolor{gray!12}
\multicolumn{8}{l}{\textit{Diffusion large language model (w/o SFT)}}\\
SDAR-8B
& 49.4 & 1.0 & 17.7 & 7.7 & 11.3 & 5.0 & 23.3 \\
\midrule

\rowcolor{gray!12}
\multicolumn{8}{l}{\textit{ChemDiffAgent (ours)}}\\
SDAR-8B + SFT
& \underline{91.3} 
& 78.4 
& \underline{94.8} 
& \underline{96.9} 
& 75.5 
& \underline{98.3} 
& 89.0  \\
SDAR-8B + SFT + RL
& \textbf{91.8} {\tiny\textcolor{red}{(+0.5)}}
& \textbf{97.1} {\tiny\textcolor{red}{(+18.7)}}
& \textbf{95.8} {\tiny\textcolor{red}{(+1.0)}}
& \underline{96.9} {\tiny\textcolor{red}{(+0.0)}}
& 79.2 {\tiny\textcolor{red}{(+3.7)}}
& \underline{98.3} {\tiny\textcolor{red}{(+0.0)}}
& \textbf{91.8} {\tiny\textcolor{red}{(+2.8)}} \\
\bottomrule
\end{tabular}
\end{adjustbox}
\vspace{-1.5em}
\end{table*}

\vspace{-0.5em}
\paragraph{Baselines.}
We compare ChemDiffAgent with a broad set of baselines. We first train Qwen3-8B as a comparable-scale autoregressive baseline. 
Note that our backbone SDAR-8B is continually pretrained from Qwen3-8B and therefore shares the same parameter scale and pretraining lineage. 
Both AR and DLLM agents are under fair comparison, including the same SFT data, training epochs, context length, and tool interface.
We also evaluate SDAR-8B without SFT as the vanilla diffusion language model baseline.
Moreover, we evaluate LLMs under ReAct framework~\citep{yao2022react}, including GLM-5.1, Gemini 2.5 Flash, DeepSeek-V3.2, GPT-5.5, and Claude Opus 4.7.
We then evaluate chemistry agents, including ChemCrow~\citep{m2024augmenting} and S1-NexusAgent~\citep{team2026s1}. 


\vspace{-0.5em}\paragraph{Implementation details.}
All experiments are conducted on 8 NVIDIA A100 GPUs.
For ChemDiffAgent, we set both the block size and the number of denoising steps to 64. All baselines are equipped with the
required tools.
Additional implementation details are provided in Appendix~\ref{app:impl}.

\subsection{Benchmarking Chemistry Agents Performance}
\label{sec:exp_main}

\paragraph{Single-turn tool-grounded chemistry reasoning.}

Table~\ref{tab:single_task_family} reports single-turn performance across six chemistry task families. We highlight three observations:
(1) Under matched model scale, training data, tool budget, and inference constraints, SDAR-8B with SFT substantially outperforms its autoregressive counterpart Qwen3-8B with SFT, improving the overall score of 19.9\%. This indicates that the diffusion paradigm provides a stronger foundation for tool-integrated chemical reasoning than autoregressive.
(2) Although our RL stage is trained on multi-turn data, our method further improves the overall score from 89.0\% to 91.8\% (+2.8\%). This indicates that preference optimization over agentic trajectories transfers effectively to simpler single-turn, tool-grounded reasoning settings.
(3) ChemDiffAgent attains the best overall score of 91.8\%, surpassing both strong frontier models equipped with ReAct (e.g., GPT-5.5 and Claude Opus 4.7) and dedicated chemistry agent frameworks such as ChemCrow~\citep{m2024augmenting}, despite relying on only an 8B backbone.


\begin{wraptable}{r}{0.61\textwidth}
\centering
\vspace{-1.8em}
\caption{Success rate on multi-turn tool-use setting. Easy, Medium, Hard correspond to 2, 3--5, 6--7 tool-calling rounds.}
\label{tab:multi_round}
\footnotesize
\vspace{+1em}
\setlength{\tabcolsep}{3pt}
\renewcommand{\arraystretch}{1.05}
\begin{tabular}{lccc|c}
\toprule
Method & Easy & Med. & Hard & Total \\
\midrule
\rowcolor{gray!12}
\multicolumn{5}{l}{\textit{Strong LLMs w/ ReAct}}\\
GLM-5.1  & 15.1 & 5.7 & 1.9 & 10.0\\
Gemini 2.5 Flash  & 21.8 & 22.1 & 6.5  & 17.8 \\
DeepSeek-V3.2     & 81.6 & 53.3 & 6.5  & 46.8 \\
GPT-5.5           & 92.0 & 63.3 & 31.8 & 61.1 \\
Claude Opus 4.7   & 98.9 & 76.4 & 44.9 & 72.8 \\
\midrule
\rowcolor{gray!12}
\multicolumn{5}{l}{\textit{Chemistry agents}}\\
ChemCrow \tiny{(Claude Opus 4.7)} & 100.0 & 75.4 & 48.6 & 73.5 \\

ChemCrow \tiny{(DeepSeek-V3.2)}
& 64.5 & 36.5 & 9.9 & 35.5 \\

S1-NexusAgent\tiny{(GPT-5.5)}
& 73.6 & 60.3 & 15.0 & 50.9 \\
\midrule
\rowcolor{gray!12}
\multicolumn{5}{l}{\textit{Autoregressive 8B model}}\\
Qwen3-8B + SFT    & 74.5 & 42.8 & 20.5 & 44.0 \\
\midrule
\rowcolor{gray!12}
\multicolumn{5}{l}{\textit{Diffusion large language model (w/o SFT)}}\\
SDAR-8B           & 14.9 & 11.0 & 8.4 & 11.2 \\
\midrule
\rowcolor{gray!12}
\multicolumn{5}{l}{\textit{ChemDiffAgent (ours)}}\\
SDAR-8B + SFT
& 75.9 
& 66.8 
& 43.0 
& 62.3  \\
SDAR-8B + SFT + RL
& 78.2 {\tiny\textcolor{red}{(+2.3)}}
& 69.9 {\tiny\textcolor{red}{(+3.1)}}
& 44.9 {\tiny\textcolor{red}{(+1.9)}}
& 64.6 {\tiny\textcolor{red}{(+2.3)}} \\
\bottomrule
\vspace{-1.8em}
\end{tabular}
\end{wraptable}

\begin{figure}[htbp]
    \centering
    \includegraphics[width=\textwidth]{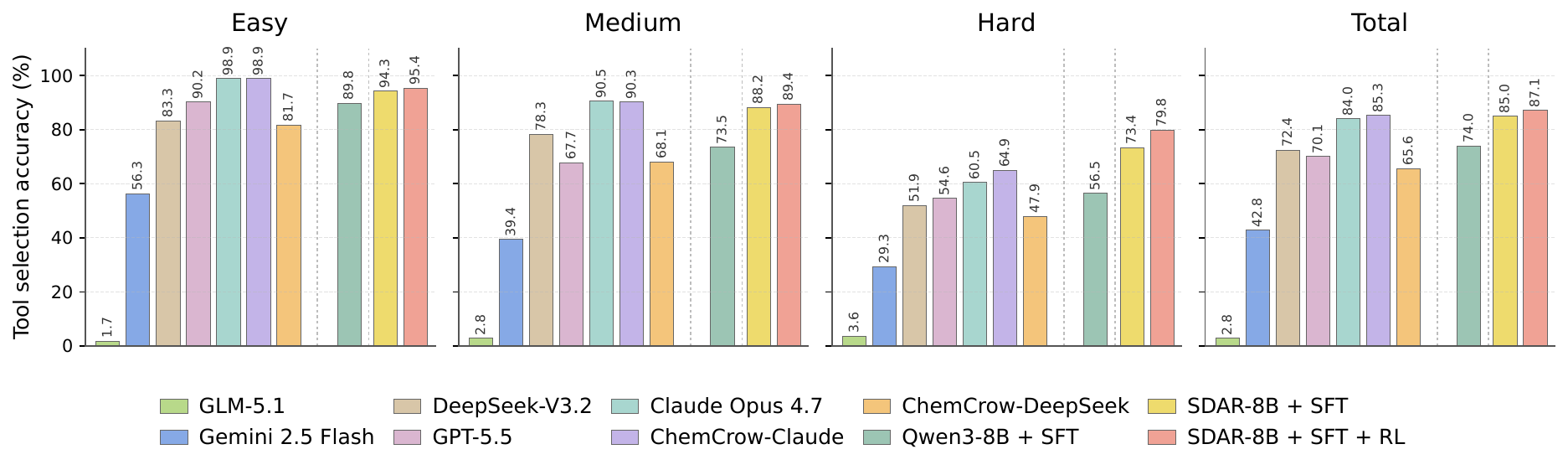}
    \vspace{-1.8em}
    \caption{Multi-turn tool selection accuracy across difficulty levels. Easy, Medium, and Hard correspond to 2, 3--5, and 6--7 tool-calling rounds, respectively.}
    \label{fig:dllm_tool}
\end{figure}

\paragraph{Multi-turn tool orchestration and reasoning.}
We then analyze multi-turn tool orchestration and reasoning performance.
Results are reported in Table~\ref{tab:multi_round}.
We observe that:
(1) Under the same SFT setting, ChemDiffAgent outperforms its autoregressive counterpart by 18.3\%, indicating that the diffusion paradigm retains a substantial advantage on long-horizon chemistry tool orchestration and reasoning.
(2) After RL training, ChemDiffAgent surpasses almost all strong frontier models including GPT 5.5, despite using only an 8B backbone.
(3) As the number of required rounds grows, all models exhibit a clear performance drop; however, ChemDiffAgent after RL remains competitive with Claude Opus 4.7 on the Hard split, confirming the strength of our approach in long-horizon tool orchestration.
We also report the tool selection accuracy across difficulty levels, reported in Fig. \ref{fig:dllm_tool}. We can see that ChemDiffAgent consistently even surpasses frontier models like GPT-5.5 and Claude Opus 4.7 in tool selection accuracy, particularly as task complexity increases.

\paragraph{Tool-use behaviour analysis.} ChemDiffAgent demonstrates more efficient and accurate tool-use. As shown in Table~\ref{tab:efficiency_analysis}, its tool-calling rounds closely align with the expected steps across various task complexities. This suggests that the agent can adaptively stop once it completes the necessary actions, rather than making redundant calls. In contrast, models like Gemini 2.5 Flash often execute more rounds than necessary, while GPT-5.5 frequently fails to invoke sufficient tool calls to solve the task.
Moreover, ChemDiffAgent maintains compact trajectory lengths, achieving the shortest single-turn output length.
These results show that ChemDiffAgent avoids long-winded reasoning or redundant tool use. Instead, it generates clear, concise, and goal-oriented trajectories, making scientific problem-solving much more efficient.

\begin{table*}[t]
\centering
\caption{Efficiency analysis of tool-use trajectories. Left: average executed tool rounds under different expected rounds. Right: average output length in single-turn and multi-turn settings.}
\vspace{+2mm}
\label{tab:efficiency_analysis}
\footnotesize
\setlength{\tabcolsep}{3pt}
\renewcommand{\arraystretch}{1.05}

\begin{minipage}[t]{0.6\textwidth}
\centering
\textbf{(a) Executed tool-calling rounds}
\smallskip

\begin{tabular}{lccccccc}
\toprule
Method & 2 & 3 & 4 & 5 & 6 & 7 & Avg. \\
\midrule
Claude Opus 4.7   & \textbf{1.99} & 3.29 & 4.16 & 5.50 & 5.65 & 6.62 & 3.90 \\
Gemini 2.5 Flash  & 5.74 & 5.81 & 5.63 & 7.00 & 6.85 & 6.36 & 6.00 \\
DeepSeek-V3.2     & 2.78 & 4.28 & 5.11 & 7.33 & 7.09 & 7.32 & 4.84 \\
GPT-5.5           & 1.90 & 2.88 & \textbf{4.05} & 6.00 & 4.76 & 5.52 & 3.42 \\
Qwen3-8B(SFT)     & 3.67 & 5.35 & 5.26 & 7.33 & 7.50 & 7.58 & 5.60 \\
\midrule
ChemDiffAgent(SFT)& 2.18 & 3.44 & 3.89 & \textbf{4.83} & \textbf{5.97} & 6.93 & 4.07 \\
ChemDiffAgent(RL) & 2.18 & \textbf{3.09} & 3.58 & \textbf{4.83} & 6.12 & \textbf{7.01} & 3.93 \\
\bottomrule
\end{tabular}
\end{minipage}
\hfill
\begin{minipage}[t]{0.36\textwidth}
\centering
\textbf{(b) Average trajectory length}
\smallskip

\begin{tabular}{lcc}
\toprule
Method & Single & Multi \\
\midrule
Claude Opus 4.7
& 456.8 & 1228.2 \\
Gemini 2.5 Flash
& 1012.8 & 2370.3 \\
DeepSeek-V3.2
& 2821.6 & 3684.2 \\
GPT-5.5
& 573.0 & 1515.4 \\
Qwen3-8B(SFT)
& 648.8 & 1769.4 \\
\midrule
ChemDiffAgent(SFT)
& 571.2 & 1746.8 \\
ChemDiffAgent(RL)
& 548.2 & 1747.7 \\
\bottomrule
\end{tabular}
\end{minipage}

\end{table*}

\subsection{Case Study}

\begin{figure}[tbp]
    \centering
    \includegraphics[width=\textwidth]{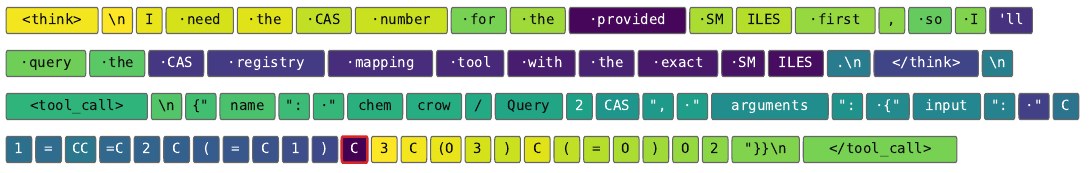}
    \vspace{-1.5em}
    \caption{Heatmap showing the relative decoding order of each token within each block of an example response. The last token of each block is highlighted with a red box.
    For brevity, only a single turn is shown; the complete response is provided in Fig~\ref{fig:denoising_case_app}.
    }
    \label{fig:denoising_case}
    \vspace{-1mm}
\end{figure}

\paragraph{Visualization of relative decoding order.} Figure~\ref{fig:denoising_case} illustrates the token-level denoising sequence of a tool-calling block in ChemDiffAgent.
Parts of the action framework, including tool-call delimiters and argument structure, are established relatively early, while selected tokens in the SMILES argument remain masked until later stages.
The resulting order interleaves structural and chemical decisions, rather than completing the argument strictly from left to right.
This behavior is desirable for tool-use agents because a valid action is not determined by the tool name alone: the tool name, argument schema, control tokens, and exact chemical input must be mutually consistent.
Therefore, ChemDiffAgent can condition low-level chemical details on a more stable global action structure, reducing premature commitment to incorrect arguments and improving the executability of tool calls.
The example thus illustrates how flexible ordering allows specific chemical decisions to be made after additional within-block context has been established.
A complete example is provided in Appendix~\ref{app:denoising_visualization}.


\paragraph{AR Agent vs. \textsc{ChemDiffAgent} Behavior Analysis.}
We compare the qualitative behaviors of AR agents and \textsc{ChemDiffAgent} in representative multi-turn chemistry tasks (see Appendix~\ref{app:case_study}). Our observations highlight three key advantages of \textsc{ChemDiffAgent}: 
\textit{\textbf{(1) Global Tool Selection:}} Unlike AR agents that repeat redundant tools due to local state extension, \textsc{ChemDiffAgent} employs global task decomposition to enable coordinated tool selection aligned with the final goal. 
\textit{\textbf{(2) Multi-source Evidence Integration:}} While AR models suffer from lossy aggregation of heterogeneous data, our model utilizes slot-level fusion to integrate diverse chemical evidence into a coherent reasoning block without detail loss. 
\textit{\textbf{(3) Robust Self-Refinement:}} \textsc{ChemDiffAgent} mitigates error propagation by validating reasoning intents and refining uncertain arguments before execution, effectively avoiding the "hallucination traps" common in AR models.

%% file: section/Conclusion.tex
\section{Conclusion and Limitations}

\paragraph{Conclusion.}
We present \textsc{ChemDiffAgent}, the first masked diffusion language agent for tool-integrated chemical reasoning that addresses the inherent limitations of AR. By reformulating reasoning as an iterative denoising process over interleaved trajectories, \textsc{ChemDiffAgent} enables bidirectional context understanding and globally coordinated planning through a specialized two-stage post-training framework. Under matched supervised fine-tuning conditions, \textsc{ChemDiffAgent} achieves gains of approximately 20 percentage points over comparable-scale AR agents, with our 8B model achieving competitive performance against frontier models such as GPT-5.5 and Claude Opus 4.7. Behavioral analyses further demonstrate more accurate tool selection and better evidence integration across multi-step workflows. These results highlight masked diffusion as a promising foundation for scientific agents that coordinate reasoning, tool use, and evidence integration, motivating further exploration of this paradigm across broader scientific domains.


\paragraph{Limitations.}
We acknowledge several limitations of the present work. First, regarding model and domain scope, our evaluation of \textsc{ChemDiffAgent} is constrained to an 8B parameter scale, a fixed toolset, and purely computational chemistry workflows. While results at this scale clearly demonstrate the advantages of the diffusion paradigm over autoregressive (AR) baselines, scaling our framework to larger backbones, broader scientific domains, and wet-lab-grounded workflows remains an important direction for future research. Second, from an infrastructure perspective, diffusion-based language models (dLLMs) still face practical deployment bottlenecks. Compared to mature AR ecosystems, serving infrastructures for dLLMs—such as efficient caching, speculative decoding, and inference optimization—remain relatively underdeveloped. Moreover, training dLLMs becomes substantially more challenging as model size and agentic reasoning horizons increase, which may make AR models more practical for current production deployments. We hope our findings provide a foundation for future work to address these system-level challenges and further unlock the potential of diffusion agents in autonomous scientific discovery.



\section*{Broader Impacts}

\textsc{ChemDiffAgent} aims to advance the development of reliable scientific reasoning agents, with the potential to accelerate chemical discovery, lower the expertise barrier for accessing computational chemistry tools, and assist researchers in tasks ranging from molecular property prediction to reaction planning. By integrating external tools rather than relying solely on parametric memory, our agent grounds its outputs in verifiable computations, which we view as a step toward more trustworthy AI assistants in scientific domains. Its ability to coordinate multiple tools may also reduce manual workflow construction and make chemistry more accessible to researchers across disciplines. Inspectable tool-use trajectories could further help researchers trace intermediate results and reproduce computational workflows. The accompanying benchmark provides a common basis for evaluating scientific tool use and may support further research into reliable diffusion-based agents.


\begin{ack}
This work is partially supported by the program of The Robotic AI-Scientist Platform of Chinese Academy of Sciences.
\end{ack}

%% file: section/Appendix.tex
\clearpage
\section{Benchmark Construction Details}
\label{app:benchmark_construction}

\paragraph{Source datasets and normalization.}
We construct our benchmark from four types of chemistry-oriented resources.
SMolInstruct~\cite{yu2024llasmol} provides instruction-style molecular tasks, ChemToolBench~\cite{wu2025chematagent} provides tool-oriented chemistry and materials science examples, MT-Mol~\cite{kim2025mt} contributes molecular optimization tasks, and T5Chem-style reaction data~\cite{lu2022unified} provides reaction prediction and synthesis-related examples.
Together, these resources cover six representative task families: composition inference, name conversion, property prediction, retrosynthesis, molecular optimization, and forward synthesis.

All raw instances are converted into a unified schema:
\begin{equation}
    z = (\tau, q, x, y, t_a),
\end{equation}
where $\tau$ is the task type, $q$ is the natural-language query, $x$ is the raw chemical input, $y$ is the reference output, and $t_a$ is the anchor tool.
The anchor tool is the minimal chemistry tool required by the original task.
For example, a molecular formula task is associated with a composition-related tool, a name conversion task is associated with a molecule identifier or canonicalization tool, and a reaction prediction task is associated with a forward synthesis or reaction prediction tool.
This anchor tool is later used to verify whether a generated trajectory remains faithful to the original seed.

\paragraph{Single-turn split.}
The single-turn split is constructed by directly converting each normalized seed into a tool-use query.
We execute a ReAct-style rollout~\cite{yao2022react} under a restricted tool environment where only the anchor tool is available.
This setting prevents the model from solving the task through unrelated tools and makes the expected tool path unambiguous.
A sample is retained if and only if the rollout successfully invokes the anchor tool and the final answer is consistent with the reference output.

We use task-specific equivalence rules rather than raw string matching whenever possible.
For molecular structures, we canonicalize SMILES before comparison.
For molecular formulas, we normalize element counts before matching.
For scalar descriptors, we allow a small numerical tolerance.
For classification-style predictions, we normalize labels before comparison.
These rules reduce false negatives caused by superficial formatting differences while still requiring the result to be chemically equivalent.
The retained examples form an execution-grounded single-tool evaluation split.

\paragraph{Chemistry tool dependency graph.}
For multi-turn construction, we first build a chemistry tool dependency graph inspired by SciToolAgent~\cite{ding2025scitoolagent}.
Each node represents a chemistry tool, and an edge $t_i \rightarrow t_j$ indicates that the output of tool $t_i$ can serve as an input or useful intermediate representation for tool $t_j$.
For example, a name resolution tool may produce a canonical SMILES string that can be consumed by property prediction, molecular optimization, or reaction analysis tools.
Similarly, a reaction prediction tool may produce products that can be further analyzed by molecular property or constraint-checking tools.

Given a seed instance $z=(\tau,q,x,y,t_a)$, we sample an executable subgraph $\mathcal{G}_s$ from the dependency graph.
The sampled subgraph must contain the anchor tool $t_a$.
During sampling, we avoid degenerate single-chain expansions when possible and encourage subgraphs with branching or converging structures.
This design produces tasks that require agents to coordinate multiple tools and aggregate evidence from different computational branches.

\paragraph{Shortcut filtering and minimal executable dependency graphs.}
We construct the ``gold'' tool-calling graph as a \textbf{minimal executable dependency graph}, rather than an arbitrary reference trajectory. We abstract shortcut filtering as a graph-search problem. For each task, we specify a starting node, the required terminal nodes, and the intermediate nodes required to complete the task. Given a sampled executable tool-use path containing $m$ tool calls, we check whether the dependency graph contains a valid path with fewer than $m$ calls that still covers all required terminal and intermediate nodes.

Specifically, starting from the initial node, we perform a bounded breadth-first search (BFS) over depths $1$ to $m-1$. This enumerates the executable tool-use paths that can be completed with fewer than $m$ calls. We then check whether any such path contains all predefined required terminal and intermediate nodes. If such a path exists, we consider it a shortcut and rewrite or remove the corresponding task; otherwise, the original task is retained. Since our tasks contain at most seven tool calls, this bounded BFS can be performed efficiently. This filtering makes the stored gold graph a minimal executable dependency structure and reduces ambiguity from redundant alternative trajectories.

\paragraph{Graph-conditioned query rewriting.}
After sampling $\mathcal{G}_s$, we use a strong teacher model to rewrite the original query into a graph-conditioned chemistry-agent task.
The teacher receives the original seed input--output pair and the sampled tool subgraph.
It is instructed to satisfy three constraints.
First, the rewritten query must preserve the semantic core of the original seed task.
Second, solving the rewritten query should require the tools in $\mathcal{G}_s$, rather than merely mentioning them.
Third, the final answer should remain grounded in the original reference output through the anchor tool.

This rewriting step turns static chemistry examples into executable multi-step agent tasks.
For molecule-centered seeds, the generated task may require resolving molecular identifiers, computing complementary properties, comparing optimized candidates, and synthesizing a final decision.
For reaction-centered seeds, the generated task may require predicting reaction outcomes, analyzing products, checking constraints, and producing a consolidated conclusion.
Thus, the generated tasks require intermediate representation conversion, multi-step chemical reasoning, and tool-result aggregation.

\paragraph{Execution verification.}
We verify each generated multi-turn task by executing it with a ReAct-style rollout in a restricted environment containing only the tools in $\mathcal{G}_s$.
The restricted tool set reduces interference from irrelevant tools and ensures that a valid trajectory must follow the sampled dependency structure.
A candidate is retained only if it satisfies all of the following conditions:
\begin{enumerate}
    \item the anchor tool $t_a$ is invoked;
    \item the runtime input to $t_a$ matches the original raw input $x$ after task-specific normalization;
    \item the output of $t_a$ matches the original reference output $y$ under the corresponding equivalence rule;
    \item the final answer is consistent with the verified tool evidence.
\end{enumerate}

Candidates that fail any of these checks are discarded.
This filtering step is important because a rewritten query may otherwise drift away from the original seed, use the sampled tools only superficially, or produce a trajectory that is not executable in the tool environment.

\paragraph{Resulting benchmark.}
The resulting benchmark contains two complementary settings.
The single-turn split evaluates whether an agent can select and execute the correct chemistry tool for a canonical task.
The multi-turn split evaluates whether an agent can coordinate multiple interdependent tools, track intermediate states, and synthesize a final answer.
Each retained instance contains an execution-verified final answer.
For multi-turn examples, we additionally store the expected calling chain or tool tree, which supports evaluation of tool selection, intermediate execution faithfulness, and long-horizon tool-use behavior.

\section{Evaluation Metrics}
\label{app:metrics}

\paragraph{Answer correctness.}
We evaluate final-answer correctness using GPT-4o as the judge.
Given a query $Q$, a reference answer $A^\star$, and a model prediction $\hat{A}$, the judge determines whether $\hat{A}$ correctly solves the task with respect to $A^\star$.
The answer correctness score for the $i$-th example is defined as
\begin{equation}
    s^{\text{ans}}_i =
    \mathbb{I}\big[
        \mathrm{Judge}(Q_i, A_i^\star, \hat{A}_i) = \mathrm{correct}
    \big],
\end{equation}
where $\mathbb{I}[\cdot]$ is the indicator function.
The final answer accuracy over a dataset $\mathcal{D}$ is computed as
\begin{equation}
    \mathrm{Acc}_{\text{ans}}
    =
    \frac{1}{|\mathcal{D}|}
    \sum_{i=1}^{|\mathcal{D}|}
    s^{\text{ans}}_i .
\end{equation}

\paragraph{Tool selection accuracy.}
Each example is associated with a gold calling chain or tool-calling sequence tree constructed as a minimal executable dependency graph through the shortcut filtering described in Appendix~\ref{app:benchmark_construction}. Thus, the gold graph is not an arbitrary reference trajectory: an alternative route to the same final answer would normally involve redundant tool calls under the benchmark construction. We extract the expected tool set from this minimal gold structure and compare it with the tools invoked by the model.
Let
\begin{equation}
    \mathcal{T}_i^\star = \{t_{i,1}^\star, t_{i,2}^\star, \ldots, t_{i,m_i}^\star\}
\end{equation}
denote the set of $m_i$ expected tools for the $i$-th example, and let $\hat{\mathcal{T}}_i$ denote the set of tools invoked by the model during its rollout.
The tool selection score for this example is computed as
\begin{equation}
    s^{\text{tool}}_i
    =
    \frac{1}{m_i}
    \sum_{j=1}^{m_i}
    \mathbb{I}\big[t_{i,j}^\star \in \hat{\mathcal{T}}_i\big].
\end{equation}
This metric only checks whether each expected tool name appears in the model trajectory at least once.
It does not evaluate tool arguments, execution outputs, or the exact order of tool calls. Answer correctness remains the primary success criterion. If an agent nevertheless discovers a correct executable solution with fewer tool calls than the stored gold structure, the answer is still considered valid and the shorter trajectory is considered more efficient; the tool-selection score is used only as a complementary diagnostic of whether the expected dependencies were covered.

The overall tool selection accuracy is the average score over all examples:
\begin{equation}
    \mathrm{Acc}_{\text{tool}}
    =
    \frac{1}{|\mathcal{D}|}
    \sum_{i=1}^{|\mathcal{D}|}
    s^{\text{tool}}_i .
\end{equation}

\paragraph{Grouped evaluation by trajectory length.}
For multi-turn analysis, we group examples according to the length of the gold calling chain.
In the main experiments, Easy, Medium, and Hard correspond to 2, 3--5, and 6--7 expected tool-calling rounds, respectively.
For each group $\mathcal{D}_g \subset \mathcal{D}$, we compute answer accuracy and tool selection accuracy as
\begin{equation}
    \mathrm{Acc}_{\text{ans}}(g)
    =
    \frac{1}{|\mathcal{D}_g|}
    \sum_{i \in \mathcal{D}_g}
    s^{\text{ans}}_i ,
\end{equation}
and
\begin{equation}
    \mathrm{Acc}_{\text{tool}}(g)
    =
    \frac{1}{|\mathcal{D}_g|}
    \sum_{i \in \mathcal{D}_g}
    s^{\text{tool}}_i .
\end{equation}
The total score is computed over all evaluated examples rather than by taking an unweighted average over groups.

\section{Implementation Details}
\label{app:impl}

All experiments are conducted on 8 NVIDIA A100 GPUs.
ChemDiffAgent uses SDAR-8B as the diffusion language model backbone.
For the main inference setting, we set the block size and the number of denoising steps to $64$.
The maximum number of tool-calling rounds is set to $7$ unless otherwise specified.

For Agentic SFT, we train SDAR-8B on the constructed chemistry-agent trajectories using the Agentic ELBO objective.
Only the agent-generated \texttt{think} and \texttt{tool\_call} regions contribute to the training loss, while external \texttt{tool\_response} regions are excluded.
We train with AdamW~\citep{loshchilov2017decoupled} using learning rate $1\times10^{-5}$.
The maximum prompt length is set to $4096$, and the maximum generation length is set to $8192$.

For Agentic VRPO, we construct preference pairs from failed trajectories generated by the SFT model and corresponding correct trajectories verified by tool execution.
We use the Agentic ELBO to estimate trajectory likelihoods for both the policy and reference models.
The preference scaling coefficient is set to $\beta=0.1$, and training uses AdamW with learning rate $5\times10^{-7}$.
For likelihood estimation, we use $2$ Monte Carlo samples.
The maximum prompt length, completion length, and total sequence length are set to $3072$, $2560$, and $6144$, respectively.

For evaluation, all non-agent baselines are run under the same ReAct-style tool-use protocol with access to the required chemistry tools.
Open-source chemistry agents are evaluated with their original execution settings.
Answer correctness is judged by GPT-4o, and tool selection accuracy is computed from the expected tool-calling sequence as described in Appendix~\ref{app:metrics}.

\section{Theoretical Analysis of Flexible Ordering}
\label{app:flexible_ordering}

We provide a detailed theoretical analysis of the advantage of flexible
generation order in DLLMs over fixed autoregressive (AR) decoding.
In structured reasoning, different subproblems can have different conditional
difficulties and downstream risks, and resolving one part may provide useful
information for another. Flexible ordering allows a DLLM to exploit these
dependencies by choosing how partially resolved states are traversed, rather
than committing to a predefined left-to-right sequence. We first show that
the BDLM objective provides training support for block-level infilling. We
then analyze two complementary advantages of flexible decoding: an adaptive
path advantage under non-uniform subproblem difficulty and a strict ordering
advantage under non-uniform decision risk.

\subsection{Block-Level Infilling Support}
\label{app:block_infilling}

We first establish that the BDLM training objective explicitly provides
learning signals over different block-level mask configurations. This provides
the training basis for resolving different parts of the output under partially
observed completion states.

\paragraph{Derivation of Eq.~\ref{eq:bdlm_block_loss}.}
Let the clean completion $y$ be partitioned into $K$ contiguous blocks
$[y^1,\ldots,y^K]$, where each block
$y^k=(y^{k,1},\ldots,y^{k,B})$ has length $B$ and $L=KB$.
For a non-empty block mask set $\emptyset\neq A\subseteq[K]$, let $y[A]$
denote the corrupted completion obtained by replacing all tokens in the
blocks indexed by $A$ with the mask token $\mathrm{M}$.

Consider a discrete block-masking process that first samples the number of
masked blocks $m\in\{1,\ldots,K\}$ and then uniformly samples a subset
$A\subseteq[K]$ with $|A|=m$. Conditioned on $m$, each block mask set has
probability
\begin{equation}
    \tilde q(A\mid |A|=m)
    =
    \frac{1}{\binom{K}{m}}.
\end{equation}
For a fixed corrupted completion $y[A]$, the reconstruction loss averages
over the tokens in the masked blocks:
\begin{equation}
    \frac{1}{|A|B}
    \sum_{k\in A}\sum_{i=1}^{B}
    \log p_\theta
    \left(
    y^{k,i}
    \mid
    y[A], y^{<k}, x
    \right).
\end{equation}

Expanding the expectation over the block-masking process gives
\begin{equation}
\begin{aligned}
\mathcal{L}^{\mathrm{block}}_\theta
=
\mathbb{E}_{(x,y)\sim p_{\mathrm{data}}}
\Bigg[
\sum_{m=1}^{K}
\frac{1}{K}
\sum_{\substack{A\subseteq[K]\\|A|=m}}
\frac{1}{\binom{K}{m}}
\frac{1}{mB}
\sum_{k\in A}\sum_{i=1}^{B}
\log p_\theta
\left(
y^{k,i}
\mid
y[A],y^{<k},x
\right)
\Bigg].
\end{aligned}
\end{equation}
Since $m=|A|$, this can be equivalently written as
\begin{equation}
\begin{aligned}
\mathcal{L}^{\mathrm{block}}_\theta
=
\mathbb{E}_{(x,y)\sim p_{\mathrm{data}}}
\Bigg[
\sum_{\emptyset\neq A\subseteq[K]}
\frac{1}{K}
\frac{1}{\binom{K}{|A|}}
\frac{1}{|A|B}
\sum_{k\in A}\sum_{i=1}^{B}
\log p_\theta
\left(
y^{k,i}
\mid
y[A],y^{<k},x
\right)
\Bigg].
\end{aligned}
\end{equation}

Thus, up to the global normalization constant $1/K$, the BDLM objective is a
linear combination of reconstruction losses over all non-empty block-level
mask sets $A\subseteq[K]$. The model therefore receives supervision over a
family of partially observed completion states, providing the block-level
infilling capability required for flexible generation order at inference time.

\subsection{Adaptive Path Advantage under Non-Uniform Subproblem Difficulty}
\label{app:adaptive_path}

Structured reasoning contains subproblems with different conditional
difficulties, and resolving one part may simplify another. Let $X$ be the
input history, $Y=(Y_1,\ldots,Y_L)$ the output, $S$ the resolved positions,
and $p$ and $q_\theta$ the data and model distributions. We define the
prediction difficulty of $Y_i$ as its conditional cross-entropy:
\begin{equation}
\begin{aligned}
d_\theta(i\mid X,Y_S)
&:=
\mathbb{E}_{Y_i\sim p(\cdot\mid X,Y_S)}
\left[-\log q_\theta(Y_i\mid X,Y_S)\right] \\
&=
H(Y_i\mid X,Y_S)
+
\delta_\theta(i\mid X,Y_S),
\end{aligned}
\end{equation}
where
\begin{equation}
\delta_\theta(i\mid X,Y_S)
=
D_{\mathrm{KL}}
\left(
p(Y_i\mid X,Y_S)
\Vert
q_\theta(Y_i\mid X,Y_S)
\right).
\end{equation}
Here, $H$ is the conditional entropy and $\delta_\theta$ is the model's
approximation error.

Let $\pi=(i_1,\ldots,i_L)$ be a generation path and
$S_t=\{i_1,\ldots,i_{t-1}\}$. Its expected loss can be written as
\begin{equation}
\mathcal{L}_\theta(\pi)
=
H(Y\mid X)
+
\mathbb{E}_p
\left[
\sum_{t=1}^{L}
\delta_\theta(i_t\mid X,Y_{S_t})
\right].
\end{equation}
The entropy term $H(Y\mid X)$ is invariant to the generation order, while the
accumulated approximation error depends on the intermediate states induced by
the path. AR follows the fixed path
$\pi_{\mathrm{AR}}=(1,\ldots,L)$, whereas DLLM decoding can select from a
larger set of paths $\Pi_{\mathrm{DLLM}}$ according to the current
partially resolved state. Therefore,
\begin{equation}
\mathcal{L}_{\mathrm{DLLM}}^*
=
\inf_{\pi\in\Pi_{\mathrm{DLLM}}}
\mathcal{L}_\theta(\pi)
\leq
\mathcal{L}_\theta(\pi_{\mathrm{AR}})
=
\mathcal{L}_{\mathrm{AR}}.
\end{equation}

This gives DLLMs an adaptive path advantage under non-uniform subproblem
difficulty. When some unresolved parts are easier to predict, or when resolving
them provides useful context for harder parts, a DLLM can choose a path with
smaller accumulated approximation error by resolving these parts first.
In contrast, AR must commit to the next position in its fixed order even when
that position remains difficult under the currently available information.

\subsection{Strict Ordering Advantage under Non-Uniform Decision Risk}
\label{app:strict_ordering}

Prediction difficulty alone does not capture the fact that errors at different
positions can have different downstream consequences. We therefore further
consider non-uniform decision risk. Let $C$ denote the available context,
$Z$ an earlier high-risk decision in the fixed AR order, and $R$ a later
lower-risk decision that provides information about $Z$. Let
$w_Z,w_R>0$ denote their expected downstream error costs.

Under execution-weighted conditional log loss, the AR order
$Z\rightarrow R$ has risk
\begin{equation}
\mathcal{J}_{\mathrm{AR}}
=
w_Z H(Z\mid C)
+
w_R H(R\mid C,Z),
\end{equation}
whereas resolving $R$ before infilling $Z$ gives
\begin{equation}
\mathcal{J}_{R\rightarrow Z}
=
w_R H(R\mid C)
+
w_Z H(Z\mid C,R).
\end{equation}
Their difference is
\begin{equation}
\mathcal{J}_{\mathrm{AR}}
-
\mathcal{J}_{R\rightarrow Z}
=
(w_Z-w_R)I(Z;R\mid C).
\end{equation}
Consequently, if
\begin{equation}
I(Z;R\mid C)>0
\qquad\text{and}\qquad
w_Z>w_R,
\end{equation}
then
\begin{equation}
\mathcal{J}_{\mathrm{DLLM}}^*
\leq
\mathcal{J}_{R\rightarrow Z}
<
\mathcal{J}_{\mathrm{AR}}.
\end{equation}

Thus, flexible ordering yields a strict advantage when a later, lower-risk
decision provides information about an earlier, higher-risk decision. Rather
than committing immediately to the high-risk position, a DLLM can first
resolve the more reliable decision and then use the resulting information to
reduce uncertainty at the more consequential one.

\paragraph{Implications for chemistry reasoning.}
Tool use and molecular generation involve decisions with different
conditional difficulties and downstream consequences.
Flexible ordering allows these differences to influence the order of
token resolution within each block, while preserving the sequential
interaction between the agent and its tools.

For tool use, some components of a call may be directly supported by
the prompt or previous observations, whereas others require further
reasoning. For example, an available molecular identifier or an explicit
property constraint may be easier to determine than an uncertain tool
choice or parameter. When these components occur in the same block,
a DLLM can resolve informative content first and use it as context for
the remaining positions. This illustrates the adaptive path advantage:
different intermediate contexts can lead to different accumulated
prediction errors. The consequences of errors also vary. A mistaken
tool name, molecular argument, or consequential parameter can change
the computation performed and affect subsequent reasoning.
When lower-risk content provides information about such a decision,
resolving it first can reduce uncertainty before the call is finalized
and executed.

For SMILES generation, local token choices participate in structural
constraints involving atom identities, bonds, branches, and ring
closures. Some positions may be strongly constrained by a reference
molecule or the requested modification, while others admit several
plausible choices. Within a block, flexible ordering can first resolve
well-supported structural elements and then use the resulting partial
structure to guide uncertain positions. These decisions can also have
unequal consequences: an incorrect bond or ring-closure token may
invalidate the representation or alter molecular connectivity, while
other choices may have a smaller effect on the task objective.
Deferring a consequential position is therefore potentially useful
when other resolvable positions provide relevant information.

The two analyses capture complementary benefits of this flexibility.
The adaptive path analysis concerns how intermediate contexts affect
model approximation error, whereas the risk-weighted analysis concerns
where information is most valuable given unequal error costs.
An easier decision is not necessarily a lower-risk decision; a useful
ordering must account for both its informational contribution and its
consequences. Together, these analyses motivate flexible within-block
resolution and establish conditions for improvements in prediction
loss and, under Bayes-optimal conditional predictions, risk-weighted
log loss.
\begin{figure}[t]
    \centering
    \includegraphics[width=\textwidth]{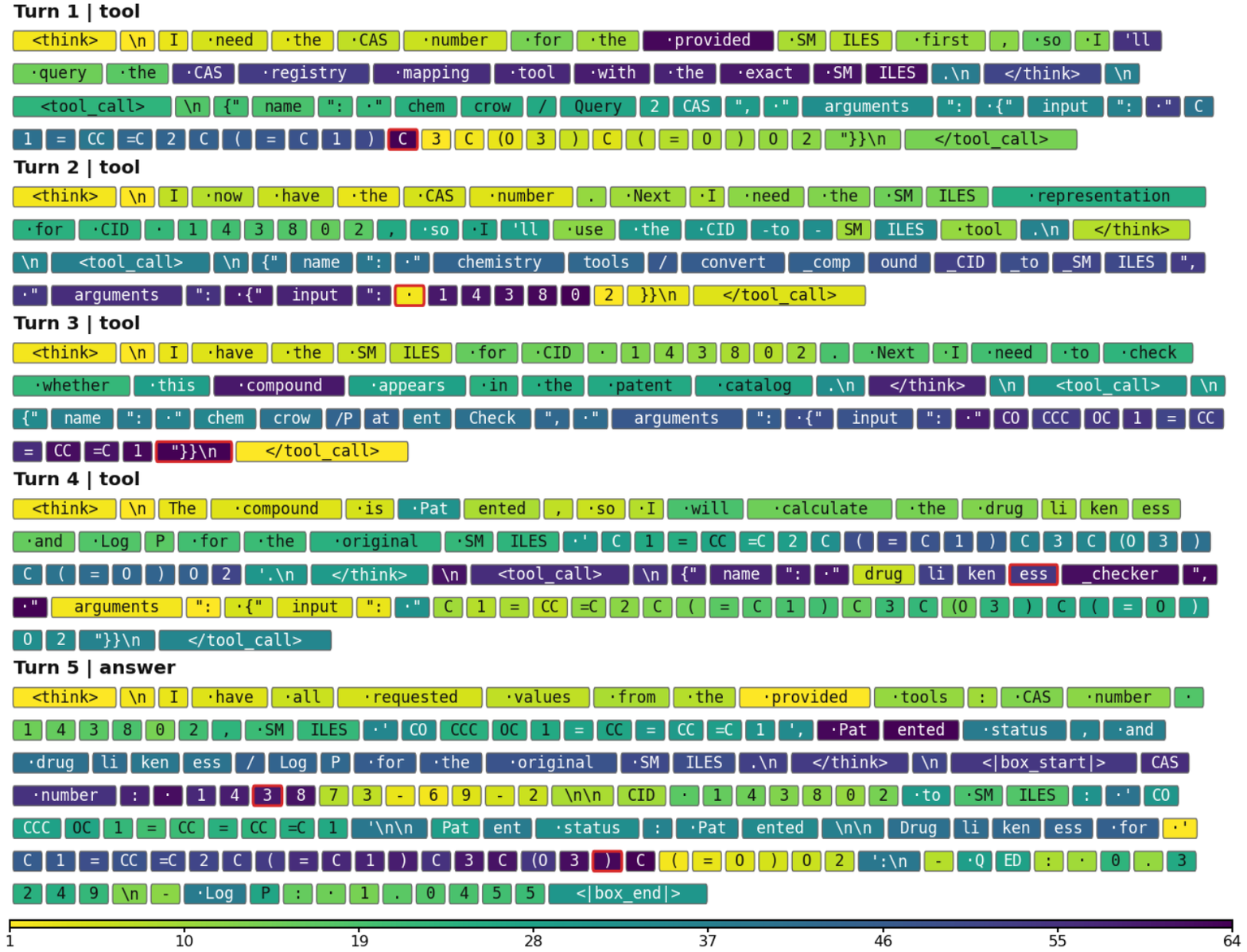}
    \caption{
    Token-level denoising order of a representative ChemDiffAgent tool-calling block.
    Lighter colors indicate tokens resolved earlier, while darker colors indicate tokens resolved later.
    }
    \label{fig:denoising_case_app}
\end{figure}

\section{Denoising Visualization of Tool-use Decisions}
\label{app:denoising_visualization}

\paragraph{Interpreting the denoising order.}
Figure~\ref{fig:denoising_case_app} shows a representative tool-calling block generated by ChemDiffAgent.
Unlike autoregressive models that commit to tokens strictly from left to right, ChemDiffAgent generates the whole block through iterative denoising.
The color of each token indicates the relative step at which the token is resolved within the block: lighter tokens are determined earlier, while darker tokens are determined later.
This allows us to inspect not only the final generated text, but also the temporal structure of the model's decision process.

The visualization reveals a non-left-to-right generation pattern. Parts of the action structure, such as tool-call delimiters and argument formatting, are resolved relatively early, while certain tokens within the SMILES argument are resolved at later stages. It reflects the flexibility of masked diffusion decoding, which allows individual token predictions to be conditioned on information resolved elsewhere within the same block. Consequently, ChemDiffAgent can determine interdependent components of a tool call in a flexible order instead of committing to each token sequentially.


\paragraph{Why this behavior is beneficial for chemistry-agent tasks.}
Tool use in chemistry agents is not merely a matter of selecting a tool name.
A valid action must jointly satisfy multiple constraints: the control token must indicate an external action, the tool namespace must match the executor, the argument schema must be valid, and the chemical string must be exact.
For example, a CAS-query tool call requires not only selecting the correct CAS lookup tool, but also passing a valid molecular representation in the expected argument field.
If any part of this structured action is inconsistent, the tool call may fail even when the model has selected the semantically correct tool.

The denoising pattern in Figure~\ref{fig:denoising_case_app} suggests that ChemDiffAgent handles this structured decision in a natural way.
By resolving the action scaffold early, the model establishes a stable context for the entire tool call.
By resolving some argument tokens later, it avoids committing too early to uncertain contents before the surrounding action format is fixed.
This is particularly useful for chemical strings such as SMILES, where small token-level mistakes can change the molecule or make the input invalid.
Thus, the model can condition local argument refinement on the already-formed global tool-use structure, improving the chance that the final action is executable.

\paragraph{Comparison with autoregressive generation.}
Autoregressive agents generate tool calls token by token.
Once an AR model emits a tool name, an opening argument field, or a partial chemical string, subsequent tokens must be generated conditioned on these previous commitments.
This left-to-right commitment can be problematic for tool-use agents because different parts of an action are mutually dependent.
The correct tool may depend on the intended argument type, the argument may depend on the selected tool schema, and the stopping or control decision depends on whether the action is already complete.
However, AR generation forces these decisions to be serialized.

ChemDiffAgent instead performs block-level generation.
Within a block, the tool name, control tokens, argument schema, and chemical string can be optimized together through denoising.
This makes the action less like a rigid left-to-right string and more like a jointly formed structured object.
The model can first settle high-confidence structural tokens and then refine lower-confidence argument tokens while still conditioning on the surrounding context.
This delayed refinement reduces premature commitment and provides a mechanism for aligning the tool-use decision with the model's current knowledge state.

\paragraph{Connection to decision boundary alignment.}
This visualization supports our view that agent behavior is a decision-making process rather than simple tool calling.
An agent must decide when internal reasoning is sufficient, when external tools are needed, which tool should be used, and how to construct a valid executable action.
In this sense, the model must align its decision boundary---when and how to act---with its knowledge boundary---what information it can reliably infer internally.
The observed denoising order is consistent with this interpretation.
ChemDiffAgent first forms the action-level structure and then refines uncertain chemical details, suggesting that external actions are committed only after a stable tool-use context has emerged.
This block-level refinement provides a structural advantage over autoregressive generation for chemistry-agent tasks, where successful execution depends on the consistency between reasoning intent, tool selection, argument format, and exact chemical input.

\raggedbottom
\section{Case Studies}
\label{app:case_study}

We provide representative case studies to illustrate qualitative differences between autoregressive agents and ChemDiffAgent.
These cases focus on three common failure modes in chemistry-agent tasks: tool selection, information aggregation, and error correction.

\vspace{1.0em}
\begin{tcolorbox}[title=\textbf{Case Study 1: Tool Selection}, colback=white, colframe=black!60]

\textbf{Problem.}
\smallskip

\emph{
Given a chemical compound name, the agent needs to first retrieve its PubChem CID, then use the CID to obtain the molecular formula, and finally compute the mass percentage of a specified element in that formula.
}

\bigskip

\begin{minipage}[t]{0.48\textwidth}
\textbf{AR Agent}

\medskip

\textit{Reasoning Style: Local Tool Continuation}

\medskip

\begin{itemize}[leftmargin=*, itemsep=3pt]
\item Starts from the compound name and selects a lookup-style tool.
\item After obtaining an intermediate identifier \(X\), it continues to call a similar name/CID lookup tool.
\item Repeatedly searches around the same identifier instead of switching to the formula-conversion tool.
\item Fails to enter the required computation stage:
    
    Compound Name 
    \(\rightarrow\) CID \(X\) 
    \(\nrightarrow\) Molecular Formula \(Y\) 
    \(\nrightarrow\) Element Percentage \(Z\)
\item The trajectory terminates with incomplete or irrelevant intermediate information.
\end{itemize}

\end{minipage}
\hfill
\begin{minipage}[t]{0.48\textwidth}
\textbf{DLLM Agent}

\medskip

\textit{Reasoning Style: Global Task Decomposition Before Execution}

\medskip

\begin{itemize}[leftmargin=*, itemsep=3pt]
\item First identifies that the question contains three distinct tool stages:
    \begin{itemize}
    \item entity resolution
    \item identifier-to-formula conversion
    \item formula-based numerical computation
    \end{itemize}
\item Builds a global dependency plan:
    
    Compound Name 
    \(\rightarrow\) CID \(X\) 
    \(\rightarrow\) Molecular Formula \(Y\) 
    \(\rightarrow\) Element Percentage \(Z\)
\item Selects a different tool for each semantic subtask.
\item Avoids repeatedly querying the same identifier-level tool.
\item Produces the final answer only after all required variables are available.
\end{itemize}

\end{minipage}

\bigskip

\textbf{Why AR Fails.}
The AR agent makes tool decisions based mainly on the latest visible state. 
After obtaining \(X\), it remains trapped in the entity-lookup stage and keeps selecting locally similar tools, rather than globally recognizing that the next required operation is formula conversion followed by numerical calculation.

\end{tcolorbox}

\clearpage
\flushbottom
\begin{tcolorbox}[title=\textbf{Case Study 2: Information Summarization}, colback=white, colframe=black!60]

\textbf{Problem.}
\smallskip

\emph{
For a molecule represented by a SMILES string, the agent needs to collect several heterogeneous properties, including predicted biological activity, solubility, molecular descriptors, functional groups, and a long side-effect profile, and return them as a single structured JSON object with fixed keys.
}

\bigskip

\begin{minipage}[t]{0.48\textwidth}
\textbf{AR Agent}

\medskip

\textit{Reasoning Style: Sequential Accumulation with Fragile Memory}

\medskip

\begin{itemize}[leftmargin=*, itemsep=3pt]
\item Queries property \(A\), then property \(B\), then property \(C\) in sequence.
\item Correctly obtains several intermediate results:
    
    \(A, B, C, D, \ldots\)
\item Encounters a verbose tool response for side effects, containing many category-probability pairs.
\item During final assembly, compresses the long response too aggressively.
\item Some values are converted into inconsistent formats:
    \begin{itemize}
    \item textual Yes/No labels become booleans
    \item numeric strings become rounded floats
    \item one predicted probability summary is partially omitted
    \end{itemize}
\item The final JSON is structurally plausible but does not faithfully preserve all required information.
\end{itemize}

\end{minipage}
\hfill
\begin{minipage}[t]{0.48\textwidth}
\textbf{DLLM Agent}

\medskip

\textit{Reasoning Style: Latent Slot-Level Aggregation}

\medskip

\begin{itemize}[leftmargin=*, itemsep=3pt]
\item Before answering, constructs a slot table for the required output keys:
    
    \[
    \{k_1, k_2, k_3, \ldots, k_n\}
    \]
\item Associates each queried tool output with a specific slot.
\item Treats the long side-effect profile as a structured field rather than as disposable explanation text.
\item Maintains type and key consistency across the final answer:
    \begin{itemize}
    \item activity labels remain aligned with the expected schema
    \item descriptor dictionaries remain nested
    \item verbose side-effect information is preserved
    \end{itemize}
\item Produces a complete JSON object after all slots have been filled.
\end{itemize}

\end{minipage}

\bigskip

\textbf{Why AR Fails.}
The AR agent can collect information step by step, but its final answer depends heavily on the last few visible states. 
When the output contains many heterogeneous fields and a long textual profile, AR is prone to lossy summarization, schema drift, and value-type mismatch. 
DLLM performs better because it maintains a global latent representation of the required output slots.

\end{tcolorbox}

\begin{tcolorbox}[title=\textbf{Case Study 3: Error Correction}, colback=white, colframe=black!60]

\textbf{Problem.}
\smallskip

\emph{
Given a reaction description, the agent must predict the major product, verify or canonicalize the generated product representation, and then compute several downstream molecular properties based on the corrected product.
}

\bigskip

\begin{minipage}[t]{0.48\textwidth}
\textbf{AR Agent}

\medskip

\textit{Reasoning Style: Commit Early and Continue}

\medskip

\begin{itemize}[leftmargin=*, itemsep=3pt]
\item Predicts a candidate product \(P_0\) from the reaction.
\item Treats \(P_0\) as reliable and immediately uses it for downstream property tools.
\item Some downstream calls return unstable or inconsistent behavior because \(P_0\) is noisy.
\item Instead of revising the product, the agent retries similar property queries.
\item The trajectory becomes:
    
    Reaction 
    \(\rightarrow P_0\) 
    \(\rightarrow\) Property Query 
    \(\rightarrow\) Retry 
    \(\rightarrow\) Retry
\item Final answer is based on an uncorrected product representation.
\end{itemize}

\end{minipage}
\hfill
\begin{minipage}[t]{0.48\textwidth}
\textbf{DLLM Agent}

\medskip

\textit{Reasoning Style: Prediction--Validation--Correction}

\medskip

\begin{itemize}[leftmargin=*, itemsep=3pt]
\item Predicts an initial product candidate \(P_0\).
\item Before computing downstream properties, checks whether \(P_0\) should be validated or canonicalized.
\item Detects that the intermediate product representation may be unreliable.
\item Applies a correction step:
    
    \[
    P_0 \rightarrow P_1
    \]
    
    where \(P_1\) is the validated/canonical product.
\item Uses \(P_1\), not \(P_0\), for all subsequent property computations.
\item Produces a final answer consistent with the corrected molecular representation.
\end{itemize}

\end{minipage}

\bigskip

\textbf{Why AR Fails.}
The AR agent commits to the first generated intermediate result and lacks a strong mechanism for revising earlier choices. 
Once the wrong or noisy product \(P_0\) enters the context, later steps inherit the error. 
DLLM is better in this case because its global planning allows an explicit correction stage between product prediction and downstream property computation.

\end{tcolorbox}

\paragraph{Summary.}
Across the three cases, we observe that the main difference between AR agents and DLLM agents lies in how they organize intermediate decisions.
AR agents tend to extend the trajectory from the most recent visible state, which makes them effective for simple local continuation but vulnerable to tool-selection traps, lossy aggregation, and early error propagation.
In contrast, DLLM agents can form a more global latent plan before committing to individual actions.
This allows them to coordinate tool dependencies, preserve heterogeneous information in structured outputs, and insert validation or correction steps before downstream computation.
These qualitative observations are consistent with our quantitative results: ChemDiffAgent achieves stronger tool selection accuracy and more stable multi-turn performance, especially in tasks requiring long-range coordination across multiple tools.

%% file: checklist.tex
\section*{NeurIPS Paper Checklist}

\begin{enumerate}

\item {\bf Claims}
    \item[] Question: Do the main claims made in the abstract and introduction accurately reflect the paper's contributions and scope?
    \item[] Answer: \answerYes{} 
    \item[] Justification: The abstract and introduction clearly state the claims made, including the contributions made in the paper and important assumptions.

\item {\bf Limitations}
    \item[] Question: Does the paper discuss the limitations of the work performed by the authors?
    \item[] Answer: \answerYes{} 
    \item[] Justification: See Appendix.

\item {\bf Theory assumptions and proofs}
    \item[] Question: For each theoretical result, does the paper provide the full set of assumptions and a complete (and correct) proof?
    \item[] Answer: \answerYes{} 
    \item[] Justification: See Appendix.

    \item {\bf Experimental result reproducibility}
    \item[] Question: Does the paper fully disclose all the information needed to reproduce the main experimental results of the paper to the extent that it affects the main claims and/or conclusions of the paper (regardless of whether the code and data are provided or not)?
    \item[] Answer: \answerYes{} 
    \item[] Justification: See section~\ref{sec:exp_setup}.

\item {\bf Open access to data and code}
    \item[] Question: Does the paper provide open access to the data and code, with sufficient instructions to faithfully reproduce the main experimental results, as described in supplemental material?
    \item[] Answer: \answerNo{} 
    \item[] Justification: The code and pre-trained models will be made publicly available upon acceptance of the paper.

\item {\bf Experimental setting/details}
    \item[] Question: Does the paper specify all the training and test details (e.g., data splits, hyperparameters, how they were chosen, type of optimizer) necessary to understand the results?
    \item[] Answer: \answerYes{} 
    \item[] Justification: See Section \ref{sec:exp_setup}.

\item {\bf Experiment statistical significance}
    \item[] Question: Does the paper report error bars suitably and correctly defined or other appropriate information about the statistical significance of the experiments?
    \item[] Answer: \answerYes{} 
    \item[] Justification: See Section \ref{sec:experiments}.

\item {\bf Experiments compute resources}
    \item[] Question: For each experiment, does the paper provide sufficient information on the computer resources (type of compute workers, memory, time of execution) needed to reproduce the experiments?
    \item[] Answer: \answerYes{} 
    \item[] Justification: See Section \ref{sec:experiments}.
    
\item {\bf Code of ethics}
    \item[] Question: Does the research conducted in the paper conform, in every respect, with the NeurIPS Code of Ethics \url{https://neurips.cc/public/EthicsGuidelines}?
    \item[] Answer: \answerYes{} 
    \item[] Justification: The research conducted in the paper conforms, in every respect, with the NeurIPS Code of Ethics.

\item {\bf Broader impacts}
    \item[] Question: Does the paper discuss both potential positive societal impacts and negative societal impacts of the work performed?
    \item[] Answer: \answerYes{} 
    \item[] Justification: See Appendix.
    
\item {\bf Safeguards}
    \item[] Question: Does the paper describe safeguards that have been put in place for responsible release of data or models that have a high risk for misuse (e.g., pre-trained language models, image generators, or scraped datasets)?
    \item[] Answer: \answerNo{} 
    \item[] Justification: The paper poses no such risks.

\item {\bf Licenses for existing assets}
    \item[] Question: Are the creators or original owners of assets (e.g., code, data, models), used in the paper, properly credited and are the license and terms of use explicitly mentioned and properly respected?
    \item[] Answer: \answerYes{} 
    \item[] Justification: All external assets used in this work—including pretrained models, datasets, and software libraries—are properly credited in the main text or references. 

\item {\bf New assets}
    \item[] Question: Are new assets introduced in the paper well documented and is the documentation provided alongside the assets?
    \item[] Answer: \answerNA{} 
    \item[] Justification: The paper does not release new assets.

\item {\bf Crowdsourcing and research with human subjects}
    \item[] Question: For crowdsourcing experiments and research with human subjects, does the paper include the full text of instructions given to participants and screenshots, if applicable, as well as details about compensation (if any)? 
    \item[] Answer: \answerNA{} 
    \item[] Justification: The paper does not involve crowdsourcing nor research with human subjects.

\item {\bf Institutional review board (IRB) approvals or equivalent for research with human subjects}
    \item[] Question: Does the paper describe potential risks incurred by study participants, whether such risks were disclosed to the subjects, and whether Institutional Review Board (IRB) approvals (or an equivalent approval/review based on the requirements of your country or institution) were obtained?
    \item[] Answer: \answerNA{} 
    \item[] Justification: The paper does not involve crowdsourcing nor research with human subjects.

\item {\bf Declaration of LLM usage}
    \item[] Question: Does the paper describe the usage of LLMs if it is an important, original, or non-standard component of the core methods in this research? Note that if the LLM is used only for writing, editing, or formatting purposes and does \emph{not} impact the core methodology, scientific rigor, or originality of the research, declaration is not required.
    \item[] Answer: \answerNA{} 
    \item[] Justification: LLMs were not used as part of the core methodology in this research.

\end{enumerate}